# Sentiment Analysis of Code-Mixed Social Media Text (Hinglish)

Gaurav Singh

School of Computing, University of Leeds, Leeds, LS29JT, UK
`gaurav_singh337@yahoo.com`

**Abstract.** This paper discusses the results obtained for different techniques applied for performing the sentiment analysis of social media (Twitter) code-mixed text written in Hinglish. The various stages involved in performing the sentiment analysis were data consolidation, data cleaning, data transformation and modelling. Various data cleaning techniques were applied, data was cleaned in five iterations and the results of experiments conducted were noted after each iteration. Data was transformed using count vectorizer, one hot vectorizer, tf-idf vectorizer, doc2vec, word2vec and fasttext embeddings. The models were created using various machine learning algorithms such as SVM, KNN, Decision Trees, Random Forests, Naïve Bayes, Logistic Regression, and ensemble voting classifiers. The data was obtained from a task on Codalab competition website which was listed as Task:9 on the Semeval-2020 competition website. The models created were evaluated using the F1-score (macro). The best F1-score of 69.07 was achieved using ensemble voting classifier.

## 1 Introduction

"Mixing languages, also known as code- mixing, is a norm in multilingual societies. Multilingual people, who are non-native English speakers, tend to code-mix using English-based phonetic typing and the insertion of anglicisms in their main language. In addition to mixing languages at the sentence level, it is fairly common to find the code-mixing behaviour at the word level. This linguistic phenomenon poses a great challenge to conventional NLP systems, which currently rely on monolingual resources to handle the combination of multiple languages. The objective of this proposal is to bring the attention of the research community towards the task of sentiment analysis in code-mixed social media text. Specifically, we focus on the combination of English with Spanish (Spanglish) and Hindi (Hinglish), which are the 3rd and 4th most spoken languages in the world respectively." (CodaLab, 2019)

So, Hinglish text is a mixture of English and Hindi language words written in English script i.e., words from Hindi vocabulary are written using English alphabets.

Sentiment Analysis is a process of analyzing the sentiments behind a text written by a person or user. This text written by the person can be on a social media website such as Facebook, Twitter, YouTube, IMDb regarding a particular topic or an experience or on any e-commerce website such as Amazon or on the merchant's or organization's website regarding a product or service consumed by them. These sentiments can be categorized in three categories such as negative, neutral, positive or in five categories such as strongly negative, negative, neutral, positive, strongly positive and so on according to the needs and wish of the individual. Further the sentiments can also be categorized as happy, angry, sad, frustrated based on emotions of the users, or as will buy, will not buy based on whether a user is interested or not (MonkeyLearn, 2020).

## 1.1 Literature Review

It was found out by Mishra et al. (2018) on performing sentiment analysis of the code-mixed dataset consisting of Hindi and English words that a maximum F1-score of 0.58 could be achieved using the char(2,6) gram features of Tf-Idf vectors on the SVM. They also tried various variations of the Tf-Idf vectorization technique such as unigrams, uni-bigrams, uni-bi-trigrams, char(3,6) gram on SVM, MLP, Bi-LSTM, and ensemble voting classifier. F1-score of 0.57 was achieved using unigrams, uni-bigrams, uni-bi-trigrams and char (3,6) gram features on SVM and a F1-score of 0.55 was achieved using glove avg features on the SVM. MLP and Bi-LSTM scored less than an F1-score of 0.55 and Voting classifier scored F1-score of 0.55. The dataset used in this analysis contained 10995 Hin-Eng code mixed training tweets and 5525 test tweets. It was different from the dataset used for the experiments in this paper.

In the research conducted by Patra et al. (2018), it was found out that a maximum F1-score of 0.569 could be achieved on the dataset consisting of Hindi-English as code-mixed language where the training dataset consisted 12936 sentences and test dataset consisted 5525 sentences. The dataset was provided to the participants of the Sentiment Analysis of Indian Language (Code Mixed) which was a shared task at ICON-2017. The maximum F1-score was achieved using the SVM and ensemble voting classifier consisting of SVM, logistic regression and Random Forest Classifier algorithms. Glove embeddings and various combinations of the Tf-Idf n-gram vector representations were used as the vector representations for the data.

Sentiment analysis performed by Ravi and Ravi (2016) on the Hinglish dataset acquired from Facebook comments and news dataset was assessed using sensitivity, specificity, AUC and t-statistics. They performed sentiment analysis using RBFNet (Radial Basis Functional Neural Network), SVM, Decision Tree, Logistic Regression, Random Forest, MLP and Naïve Bayes algorithms. They used Tf-Idf vectorization technique for vector representations. It was found out by them that RBF Neural network performed best followed by SVM (RBF), Random Forest, SVM (Linear) and Logistic Regression for the news dataset. And for the Facebook comments dataset again RBF Neural Network performed best followed by the Logistic Regression, Random Forests, SVM (Linear), MLP and Naïve Bayes.

Experiments conducted by the Kaur et al. (2019) on the Hinglish dataset resulted in achieving the best F1 scores from SVM and the Logistic Regression. The dataset was collected from the comments section of two Cookery channels on YouTube. The dataset was classified into 7 categories, each category having equal number of samples. Tf-Idf vectorizer, count vectorizer and term frequency vectorizer were used as the vectorization techniques for converting the data.

So, it was observed that Tf-Idf vectorizer got the best results as the vectorization technique to transform the data. And ensemble voting classifier, SVM, Random Forest, Logistic Regression gave the best results for classifying the data written in Hinglish language.

## 2  Data

The data was obtained from the **Task 9: Sentiment Analysis of Code-Mixed Social Media Text (Hinglish)** of SemEval-2020 competition by being its participant. The data consisted of Code-Mixed tweets containing Hindi and English words written in English script. The tweets were classified among the Negative, Neutral or Positive sentiment polarity. A few examples of the tweets obtained after consolidating and cleaning them from the raw data are given as follows:

1. "haan yaar neha kab karega woh post usn na sach mein photoshoot karna chahiy phir woh post karega" – neutral.

2. "all india nrc lagu kare w kashmir se dhara ko khatam kare ham indian ko aps yahi umid hai" – positive.

3. "hadd hai kamin pan ki aapko pata hai Ye kev jail gaya hai Ye bina order ke railway platform pa" – negative.

The data was collected by downloading the files from the links provided separately for the training, test and validation data on the Codalab website. The files were provided in the .txt format.

### 2.1  Data Description

Data was provided in the .txt files in a tab separated format. Each tweet was split up in the corresponding words and symbols represented as tokens which were present in the separate lines. A keyword was provided along with each token which denoted whether the word belonged to English language or Hindi language or a universal symbol. Starting of each tweet was marked by providing the metadata about the tweet in the separate line before the first word of the tweet which mentioned sentiment polarity and unique identifier for that tweet. In the test data file, the sentiment polarity was not provided as the meta data for the tweets, only the unique identifier was provided for each tweet.

### 2.2  Data Exploration

After consolidating the data (tweets) into the sentences from the individual words provided in the raw data files, it was found out that the training data contained *14000* tweets, test data and validation data contained *3000* tweets each.

On exploring the data, it was found out that out of total 14000 tweets in training data *4,102* belonged to negative sentiment polarity, *4,634* belonged to positive sentiment polarity and *5,264* belonged to neutral sentiment polarity. The test and validation data also had similar distribution of the tweets among the three sentiment polarities. The following figure shows the distribution of the tweets among the sentiment polarities for the training, test and validation data:

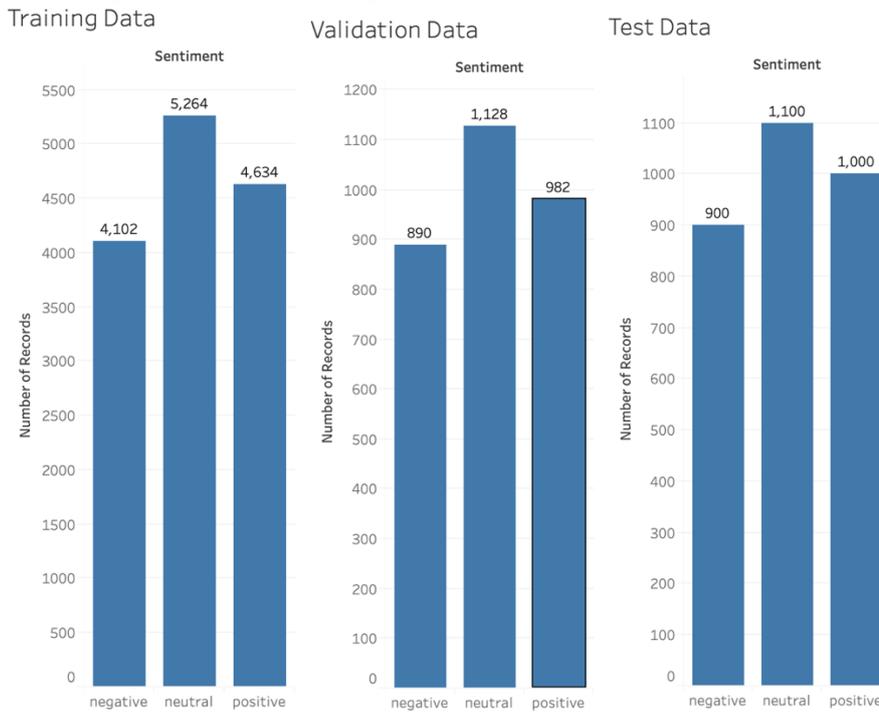

**Fig. 1.** Distribution of Training, Validation and Test Data among sentiment polarities.

On further exploration of the data, it was found out that the training data contained *169893* words tagged as Hindi and *121412* words tagged as English. On making the dictionary of their frequency distribution, it was found out that there were *26653* unique Hindi words and *26082* unique English words.

### 2.3  Data Quality Issues

A lot of data quality issues were encountered while processing the data. The first data quality issue encountered while converting the data into the sentences from the individual words given in the text files was the presence of the double inverted commas (") as the token. The data was tried to read into the pandas dataframes from the text files which resulted into the following error:

```
ParserError: field larger than field limit (131072)
```

**Fig. 2.** Error while reading the text files into pandas dataframe.

After replacing the double inverted commas with the empty string in the text files, the data was read successfully into the dataframes. The following code used for reading the data into the dataframes:

```
df = pd.read_csv("/Users/gaurav/Desktop/conll/train_14k_split_conll.txt",sep='\t',engine='python',
                 names=['word','type','class'])
```

**Fig. 3.** Code for reading the data into dataframes.

After consolidating the tweets into the sentence form, various other data quality issues were encountered which are mentioned as follows:
1. Presence of https links. Https links cannot help in determining the sentiment polarity of the tweets as these can be associated with any of the three polarities.
2. Presence of @ User tags. These tags cannot help in determining the sentiment polarity as a single user can be associated with all of the three polarities.
3. Presence of # tags. The hash tags were considered as unwanted symbols. However, the words following the hash tags were not considered as data quality issue as these can be associated with a particular sentiment polarity. So, these words were not broken up into the corresponding words.
4. Presence of emoticons.
5. Presence of RT (Retweet keyword).
6. Presence of punctuation marks, numbers and other unwanted symbols.
7. Presence of English and Hindi stopwords.
8. Spelling mistakes done by people in writing the Hindi words in English script as well as writing the English vocabulary words.
   Since there are no specified spellings for writing the Hindi words in English script, there were observed to be a lot variations in spellings of the same Hindi word when written in English. The spelling of the word written most for the same Hindi word was assumed to be the correct spelling for that Hindi word.

The last two data quality issues, presence of Hindi stopwords and spelling mistakes done by people in writing the Hindi words in the English script were specific to the task where Hindi language words were written by people in English script. So, they were assumed to play an important role for cleaning the code-mixed social media text written in Hinglish as there are no available resources to perform these tasks.

9. Presence of only https links in the tweets. So, when the https links were removed, the tweets became empty strings.
10. Incorrect tagging of words. Many Hindi words were tagged as English and many English words were tagged as Hindi. After making the dictionary from the frequency distribution of Hindi and English words, it was observed that many Hindi words were tagged more times as English and vice versa. The following two images shows this observed data quality issue:

```
In [108]: print("Frequency of 'aatankwadi' in Hindi Dictionary = ",hindi_fdist['aatankwadi'])
          print("Frequency of 'aatankwadi' in English Dictionary = ",english_fdist['aatankwadi'])

          Frequency of 'aatankwadi' in Hindi Dictionary =  1
          Frequency of 'aatankwadi' in English Dictionary =  2
```

**Fig. 4.** Wrong tagging of the Hindi word 'aatankwadi'.

```
In [7]: print("Frequency of 'anthem' in Hindi Dictionary = ",hindi_fdist['anthem'])
        print("Frequency of 'anthem' in English Dictionary = ",english_fdist['anthem'])
        Frequency of 'anthem' in Hindi Dictionary =  2
        Frequency of 'anthem' in English Dictionary =   1
```

**Fig. 5.** Wrong tagging of the English word 'anthem'.

## 3 Data Preparation

The raw data was first consolidated into the sentences from the individual words provided for each tweet. Then data was cleaned in five iterations, removing more noise from the data with every iteration and making the data more suitable for modelling. The data was then transformed into vectors using various vectorization techniques using the predefined libraries in python.

### 3.1 Data Consolidation

Raw data given in the form of words and symbols present in separate lines was consolidated to make sentences for the individual tweets after combining the words and symbols for the individual tweets. Dataframes data structure provided by the pandas library of the python was used for this purpose. After converting tweets into the corresponding sentences, they were stored along with their uid and sentiment polarity in the dataframes. This process was done for each of the training, validation and test data files. The following image shows the snippet of code used for this process.

```
In [60]: text=""
         cla=""
         cla=df['class'][0]
         j=df['type'][0]
         for i in range(1,len(df)):
             if(df['word'][i]=="meta"):
                 df1 = df1.append({"ID": j, "Text": text, "class": cla}, ignore_index=True)
                 cla=df['class'][i]
                 j=df['type'][i]
                 text=""
             else:
                 text += " " + str(df['word'][i])
         df1 = df1.append({"ID": j, "Text": text, "class": cla}, ignore_index=True)
```

**Fig. 6.** Code for reading the data from the text files.

### 3.2 Data Cleaning

Data cleaning was performed in 5 iterations, improving the data quality with each iteration for modelling and results of experiments conducted were noted after each iteration. Custom methods for cleaning the data were made for each iteration using regex and other predefined libraries in python.

### 3.2.1 Iteration 1

In this iteration, @ User tags were removed, html tags were removed, html decoding was converted to symbols and special characters and punctuation marks were removed. The following image shows the snippet of code used for this iteration.

```python
In [14]: import re
         from bs4 import BeautifulSoup
         def clean_tweet(tweet):
             text = re.sub(r'@[A-Za-z0-9]+','', tweet)  # remove mentions [like @User]
             text = BeautifulSoup(text, 'lxml').get_text() # remove html tags and converts html decoding [things like ", >
             text = re.sub("[^a-zA-Z]", " ", text) # only keeps in letter characters [removes numbers, hash tags and other special
             return text
```

**Fig. 7.** Data Cleaning iteration 1.

### 3.2.2 Iteration 2

In this iteration RT (Retweet keyword) was removed, https links were removed, nan (null character) keyword was removed in addition to the data cleaning performed in iteration 1. The following image shows the snippet of code used for this iteration.

```python
In [504]: import re
          from bs4 import BeautifulSoup
          def clean_tweet(tweet):
              text = re.sub(r'@ [A-Za-z0-9\']+','', tweet)  # Removes mentions [like @ User]
              text = BeautifulSoup(text, 'lxml').get_text() # Removes html tags and converts html decoding [things like ", >
              text = re.sub(r'https (//)[A-Za-z0-9. ]*(/) [A-Za-z0-9]+','',text) #Removes https links
              text = re.sub(r'https[A-Za-z0-9/. ]*','',text) #Removes https links
              text = re.sub("[^a-zA-Z]", " ", text) # Only keeps in letter characters [removes numbers, hash tags and other special
              text = re.sub(r'\bRT\b',' ', text)  # Removes RT (ReTweet keyword)
              text = re.sub(r'\bnan\b',' ', text)  # Removes nan (Null value keyword)
              return text
```

**Fig. 8.** Data Cleaning iteration 2.

### 3.2.3 Iteration 3

In this iteration a Hindi dictionary was created for normalizing the different spellings of the same Hindi words written. English stopwords were removed and English words were stemmed using the predefined PorterStemmer library of python. All the words were also lowercased. These functions were performed in addition to the data cleaning performed in iteration 2.

The Hindi dictionary created for normalizing the Hindi words took care of the following spelling mistakes in the Hindi words:

1. The vowels or letters used interchangeably by the people that create the same phoneme effect. People usually confuse letter *q* with *k*, letter *j* with *z*, letter *o* with *u*, letter *w* with *v*. These pairs of letters were considered as same.
2. The letters which are repeated sequentially to create an elongated phoneme effect, were considered only once.
3. The missing vowels in the words to write the words in short form were taken care of.

The following image shows a part of the vocabulary created for normalizing the Hindi words.

| | A | B | C | D | E | F | G | H | I | J | K | L |
|---|---|---|---|---|---|---|---|---|---|---|---|---|
| 1 | aa | aaa | aaaa | aia | | | | | | | | |
| 2 | aaaaaaah | aahh | ah | ahh | ahhh | ahhhh | | | | | | |
| 3 | aaafat | aafat | afaat | | | | | | | | | |
| 4 | aaag | aag | ag | agg | agggg | | | | | | | |
| 5 | aaage | aagae | aage | age | agge | | | | | | | |
| 6 | aaahirwaad | | | | | | | | | | | |
| 7 | aaaj | aaj | aaz | aj | ajj | ajjj | az | azeez | | | | |
| 8 | aaana | aana | ana | anna | annaa | annaaaa | | | | | | |
| 9 | aaane | aane | ane | | | | | | | | | |
| 10 | aaap | aap | ap | app | | | | | | | | |
| 11 | aaarakshan | aarakshan | aarkshan | | | | | | | | | |
| 12 | aaay | aay | aey | ayaaaaay | ayyy | ayyyy | | | | | | |
| 13 | aab | ab | abb | aeb | | | | | | | | |
| 14 | aabaad | abaad | abad | | | | | | | | | |
| 15 | aabaadi | aabadi | abaadi | abadi | | | | | | | | |
| 16 | aabe | abe | abee | | | | | | | | | |
| 17 | aabhari | | | | | | | | | | | |
| 18 | aabhi | abhai | abhi | abhiii | | | | | | | | |
| 19 | aabo | abbu | | | | | | | | | | |
| 20 | aaby | abay | abey | | | | | | | | | |
| 21 | aacha | acaha | accha | acchha | acha | achaa | achacha | achcha | achchaa | achchha | achha | achhha |
| 22 | aachaar | aachr | achaar | achar | | | | | | | | |
| 23 | aache | aachhe | acche | achche | achchhe | ache | achhe | | | | | |
| 24 | aachi | acchi | achai | achchi | achhi | achi | | | | | | |
| 25 | aacrew | | | | | | | | | | | |
| 26 | aadami | aadmi | admi | | | | | | | | | |
| 27 | aadaraniya | aadarniya | adaraniya | adarniya | | | | | | | | |
| 28 | aadarniy | | | | | | | | | | | |
| 29 | aadarsh | adarsh | | | | | | | | | | |
| 30 | aadarshoon | | | | | | | | | | | |
| 31 | aadat | adat | | | | | | | | | | |
| 32 | aadath | adith | | | | | | | | | | |
| 33 | aadesh | | | | | | | | | | | |
| 34 | aadh | | | | | | | | | | | |
| 35 | aadhar | adhar | | | | | | | | | | |
| 36 | aadhe | | | | | | | | | | | |
| 37 | aadi | aadiii | adi | | | | | | | | | |

**Fig. 9.** Hindi words vocabulary created for Normalization.

The first word was used for replacing the variations of the spellings of the same Hindi word. This word was the one which had the highest frequency in the dataset among all the variations of its spellings.

### 3.2.4   Iteration 4

In this iteration a custom stopword list was created. It was used for removing the Hindi stopwords from the data along with the data cleaning performed in the iteration 3. The stopwords list was created using the words in the training data which had a length ranging from 1 to 3 which were 2641 in number. Then this list was checked and words which did not act as stopwords were removed from this list which were 263 in number. The following image shows the words marked as not stopwords.

```
In [53]: not_stopwords = ['zoo','yug','yog','wtf','wrk','wrd','wqt','wow','wot','wor','won','win','wig','wht','who','whn',
          'wha','wax','wat','war','veg','usa','umr','tym','typ','txt','two','tru','trp','toy','tip','tie',
          'tez','tea','tax','tag','swt','sun','sur','sum','sue','sir','sit','six','sip','sin','sim','sex',
          'sea','run','rod','req','raw','rat','rap','pyr','pwd','pvt','pvr','put','pre','pro','prk','ppl',
          'pop','pol','pod','pmo','pm','plz','pig','phd','pic','pet','pen','peg','pay','pat','pan','pak','own',
          'owl','owe','out','oop','omg','old','oil','odd','nvr','nut','nt','nrc','not','non','no','nmo','nii',
          'ni','nip','nhn','nhi','nhh','nhe','nh','ngo','new','net','ncr','nct','nai','nah','naa','na','mza',
          'mum','muh','muj','mug','mud','mra','mop','mom','mol','moh','mod','mna','mje','mja','mix','mit','met',
          'men','mba','max','mat','mar','mad','maa','ma','luv','lut','loo','lok','log','lot','lov','low','lol',
          'imp','law','lar','lag','lad','kyu','kyo','kyn','kom','joy','jog','job','jio','jaw','jat','hug','how',
          'hot','hor','hop','hog','hit','hip','hlp','hiv','hat','hap','hag','gun','gum','grt','god','gen','gem',
          'gdp','gay','gau','gap','fut','fun','fur','fox','fod','fix','fit','fek','fed','fav','fat','far','fan',
          'fab','eye','eve','eng','end','elm','ego','egg','eat','ear','due','don','dog','die','den','del','dar',
          'dal','dad','cut','cup','cum','coy','cow','cop','con','col','chu','cho','ceo','cat','car','cap','bun',
          'bum','bug','bsd','bra','bkl','bjp','big','bhn','bhg','beg','bed','bbc','bat','bap','bad','awe','avg',
          'ate','ant','aim','aid','age','add','ada','act','ace','aba','aag']
```

**Fig. 10.** Non stopwords list.

*3.2.5 Iteration 5*

In this iteration emoticons were converted into words along with the data cleaning performed in the previous iteration. The emoticons were converted using the emoji library of the python and word 'face' was removed as the stopword from these converted words as this word does not help in determining the sentiment polarity.

**3.3 Data Transformation**

The data was transformed into vectors of numbers using several vectorization techniques using the predefined libraries in python. The vectorization techniques used were Count Vectorizer, One Hot Binarizer, Tf-Idf Vectorizer, Word2Vec embeddings, Doc2Vec Embeddings, FastText Embeddings. Scikit Learn library and Gensim library of python were used for implementing these techniques.

Various settings were explored for the Count Vectorizer, One Hot Binarizer and Tf-Idf Vectorizer for making the vectors of data. Different n-grams settings considered for making the vectors were Unigrams, Bigrams, Trigrams, Uni-Bigrams, Bi-Trigrams, Uni-Bi-Trigrams where words were chosen for making these n-grams. Different minimum occurrence frequencies of the n-grams considered for making the vectors were 1, 2, 3 and 5 using each of these n-gram settings.

# 4 Modelling

Models were created on the data obtained after each iteration of data cleaning and results were noted for the experiments conducted. Models were created using Support Vector Machines, KNN, Decision Tree Classifiers, Gaussian Naïve Bayes Classifier, Multinomial Naïve Bayes Classifier, Logistic Regression, Random Forests Classifier and ensemble voting classifier. Multi Layered Perceptron and Convolutional Neural Network were also tried but with limited number of experiments. Scikit Learn and keras library of python were used for creating the models.

Random state was set as *0* and class weight was set to *balanced* in the parameters for the SVM, Decision Tree Classifier, Logistic Regression Classifier and the Random Forest Classifier. The kernel parameter in SVM was set to *linear*. The k parameter for the KNN was experimented in the range of *0 to 100* for finding the best F1-score. The Random Forest was provided with *1000* as the parameter for the n estimators. The probability parameter for the SVM was set to *true* in the ensemble voting classifier.

**4.1 Experiment 1**

In this experiment models were created on the data obtained after iteration 1 of data cleaning. Count Vectorizer, One hot binarizer and Tf-Idf vectorizer were used as feature extraction techniques on the cleaned data. Only unigrams of words were considered keeping the minimum occurrence frequency of n-grams as 1 for creating the vectors. The F1-scores recorded for these experiments are shown in the table below.

Table 1. F1-Scores for Experiment 1.

| F1-Score | SVM | KNN | Decision Tree | Gaussian Naïve Bayes | Multinomial Naïve Bayes | Logistic Regression | Random Forest |
|---|---|---|---|---|---|---|---|
| Count Vectorizer | 0.6060 | 0.4210 | 0.5504 | 0.4917 | 0.6301 | 0.6385 | 0.6524 |
| One Hot Binarizer | 0.6073 | 0.3995 | 0.5349 | 0.4910 | 0.6319 | 0.6475 | 0.6564 |
| Tf-Idf Vectorizer | 0.6650 | 0.6326 | 0.5333 | 0.4844 | 0.6575 | **0.6654** | 0.6483 |

The experiment results and code can be viewed at https://github.com/sc19gs/Hinglish/blob/master/Experiment%201.ipynb.

### 4.2 Experiment 2

In this experiment models were created on the data obtained after iteration 2 of data cleaning. Count Vectorizer, One hot binarizer and Tf-Idf vectorizer were used as feature extraction techniques on the cleaned data. Only unigrams of words were considered keeping the minimum occurrence frequency of n-grams as 1 for creating the vectors. The F1-scores recorded for these experiments are shown in the table below.

Table 2. F1-Scores for Experiment 2.

| F1-Score | SVM | KNN | Decision Tree | Gaussian Naïve Bayes | Multinomial Naïve Bayes | Logistic Regression | Random Forest |
|---|---|---|---|---|---|---|---|
| Count Vectorizer | 0.6062 | 0.4503 | 0.5373 | 0.4945 | 0.6336 | 0.6442 | 0.6604 |
| One Hot Binarizer | 0.6145 | 0.4409 | 0.5371 | 0.4946 | 0.6342 | 0.6532 | 0.6568 |
| Tf-Idf Vectorizer | 0.6618 | 0.6418 | 0.5390 | 0.4821 | 0.6654 | **0.6669** | 0.6612 |

The experiment results and code can be viewed at https://github.com/sc19gs/Hinglish/blob/master/Experiment%202.ipynb.

### 4.3 Experiment 3

In this experiment models were created on the data obtained after iteration 3 of data cleaning. Count Vectorizer, One hot binarizer and Tf-Idf vectorizer were used as feature extraction techniques on the cleaned data. Only unigrams of words were considered keeping the minimum occurrence frequency of n-grams as 1 for creating the vectors. The F1-scores recorded for these experiments are shown in the table below.

Table 3. F1-Scores for Experiment 3.

| F1-Score | SVM | KNN | Decision Tree | Gaussian Naïve Bayes | Multinomial Naïve Bayes | Logistic Regression | Random Forest |
|---|---|---|---|---|---|---|---|
| Count Vectorizer | 0.6285 | 0.4467 | 0.5591 | 0.4530 | 0.6511 | 0.6624 | 0.6775 |
| One Hot Binarizer | 0.6278 | 0.4441 | 0.5551 | 0.4485 | 0.6512 | 0.6600 | 0.6692 |
| Tf-Idf Vectorizer | 0.6769 | 0.6581 | 0.5556 | 0.4503 | 0.6745 | **0.6798** | 0.6679 |

The experiment results and code can be viewed at https://github.com/sc19gs/Hinglish/blob/master/Experiment%203.ipynb.

A significant improvement was observed in the results from the previous experiment due to the normalization of Hindi words.

## 4.4 Experiment 4

In this experiment models were created on the data obtained after iteration 4 of data cleaning. Count Vectorizer, One hot binarizer and Tf-Idf vectorizer were used as feature extraction techniques on the cleaned data. Only unigrams of words were considered keeping their minimum occurrence frequency as 1 for creating the vectors. The F1-scores recorded for these experiments are shown in the table below.

Table 4. F1-Scores for Experiment 4.

| F1-Score | SVM | KNN | Decision Tree | Gaussian Naïve Bayes | Multinomial Naïve Bayes | Logistic Regression | Random Forest |
|---|---|---|---|---|---|---|---|
| Count Vectorizer | 0.6252 | 0.4993 | 0.5839 | 0.4549 | 0.6615 | 0.6621 | 0.6633 |
| One Hot Binarizer | 0.6277 | 0.4958 | 0.5869 | 0.4524 | 0.6541 | 0.6595 | 0.6638 |
| Tf-Idf Vectorizer | 0.6667 | 0.6458 | 0.5874 | 0.4537 | 0.6747 | **0.6794** | 0.6630 |

The experiment results and code can be viewed at https://github.com/sc19gs/Hinglish/blob/master/Experiment%204.ipynb.

The results of Logistic Regression, Random Forest and SVM declined a bit. But the results of Decision Tree, Gaussian Naïve Bayes and Multinomial Naïve Bayes saw an improvement.

## 4.5 Experiment 5

In this experiment models were created on the data obtained after iteration 5 of data cleaning. Count Vectorizer, One hot binarizer and Tf-Idf vectorizer were used as feature extraction techniques on the cleaned data. Only unigrams of words were considered keeping their minimum occurrence frequency as 1 for creating the vectors. The F1-scores recorded for these experiments are shown in the table below.

Table 5. F1-Scores for Experiment 5.

| F1-Score | SVM | KNN | Decision Tree | Gaussian Naïve Bayes | Multinomial Naïve Bayes | Logistic Regression | Random Forest |
|---|---|---|---|---|---|---|---|
| Count Vectorizer | 0.6145 | 0.4882 | 0.5891 | 0.4445 | 0.6548 | 0.6632 | **0.6784** |
| One Hot Binarizer | 0.6261 | 0.4779 | 0.5961 | 0.4429 | 0.6512 | 0.6604 | 0.6755 |
| Tf-Idf Vectorizer | 0.6636 | 0.6515 | 0.5805 | 0.4470 | 0.6719 | 0.6715 | 0.6735 |

The experiment results and code can be viewed at https://github.com/sc19gs/Hinglish/blob/master/Experiment%205.ipynb.

### 4.6 Experiment 6

In this experiment models were created on the data obtained after iteration 5 of data cleaning and was transformed using word2vec, doc2vec and fasttext word embeddings. Some models were also created using CNN and MLP architectures using word2vec and fasttext embeddings. The F1-scores recorded for these experiments are shown in the table below.

Table 6. F1-Scores for Experiment 6.

| F1-Score | SVM | KNN | Decision Tree | Gaussian Naïve Bayes | Logistic Regression | Random Forest |
|---|---|---|---|---|---|---|
| Word2Vec | 0.6505 | 0.6525 | 0.5082 | 0.5734 | 0.6597 | **0.6619** |
| Doc2Vec | 0.5276 | 0.5360 | 0.4452 | 0.4805 | 0.5229 | 0.5403 |
| FastText | 0.5651 | 0.5914 | 0.4621 | 0.4988 | 0.5645 | 0.5919 |

The following screenshot of code shows the architecture used for making the models using MLP.

```
In [341]: model = keras.Sequential()
          model.add(Dense(nr_in,activation='relu'))
          model.add(Dense(300, activation = 'relu'))
          model.add(Dense(300, activation = 'relu'))
          model.add(Dense(200, activation = 'relu'))
          model.add(Dense(200, activation = 'relu'))
          model.add(Dense(100, activation = 'relu'))
          model.add(Dense(100, activation = 'relu'))
          model.add(Dense(3,activation='softmax'))
```

**Fig. 11.** MLP Architecture.

The following screenshot of code shows the architecture used for making the models using CNN.

```
In [341]: model.add(Conv1D(8, kernel_size=(3),activation='relu',padding='same'))
          model.add(MaxPooling1D((4),padding='same'))
          model.add(Dropout(0.2))
          model.add(Conv1D(4, kernel_size=(3),activation='relu',padding='same'))
          model.add(MaxPooling1D((2 ),padding='same'))
          model.add(Dropout(0.2))
          model.add(Flatten())

          model.add(Dense(128, activation='relu'))
          model.add(Dense(3, activation='softmax'))
```

**Fig. 12.** CNN Architecture.

The following table shows the F1-scores obtained on models created using MLP and CNN architectures.

Table 7. F1-Scores for Experiment 6 on MLP and CNN.

| F1-Score | MLP | CNN |
|---|---|---|
| Word2Vec | 0.6100 | 0.5969 |
| FastText | 0.5079 | 0.4325 |

The experiment results and code can be viewed at https://github.com/sc19gs/Hinglish/blob/master/Experiment%206.ipynb.

### 4.7 Experiment 7

In this experiment different settings of the Count Vectorizer, One Hot Binarizer and Tf-Idf Vectorizer were explored by using them on data obtained after iteration 5 of data cleaning. Various n-gram settings along with their various minimum occurrence frequencies were considered for creating the vectors.

Following table shows the F1-scores obtained when uni-bi-trigrams of the tf-idf vectors were used with minimum n-gram occurrence frequencies ranging from 1 to 5 for making the models. The *f* in the following table denotes the minimum occurrence frequency of the n-gram and *KC* denotes kernel crashed due to lack of resources.

Table 8. F1-Scores for uni-bi-trigrams of tf-idf vectors with various n-gram frequencies on data obtained after iteration 5 of data cleaning.

| F1-Score | SVM | KNN | Decision Tree | Gaussian Naïve Bayes | Multinomial Naïve Bayes | Logistic Regression | Random Forest |
|---|---|---|---|---|---|---|---|
| f = 1 | 0.6689 | 0.6530 | 0.5789 | KC | KC | 0.6676 | 0.6706 |
| f = 2 | 0.6654 | 0.6350 | 0.5835 | 0.5105 | 0.6674 | **0.6800** | 0.6676 |
| f = 3 | 0.6619 | 0.6308 | 0.5674 | 0.4877 | 0.6781 | 0.6746 | 0.6692 |
| f = 5 | 0.6671 | 0.5696 | 0.5791 | 0.4786 | 0.6692 | 0.6723 | 0.6639 |

The Unigrams, Bigrams, Trigrams, Uni-Bigrams and Bi-Trigrams of the words with minimum n-gram frequencies ranging from 1 to 5 were also experimented for this data. The maximum F1-score of 0.6784 was obtained with Unigrams, maximum F1-score of 0.5646 was obtained with Bigrams, maximum F1-score of 0.3599 was obtained with Trigrams, maximum F1-score of 0.6797 was obtained with Uni-Bigrams and maximum F1-score of 0.5601 was obtained with Bi-Trigrams.

The results of experiments conducted on uni-bi-trigrams using other vectorizers, on above mentioned n-gram settings using all three vectorizers and code can be viewed at https://github.com/sc19gs/Hinglish/blob/master/Experiment%207.ipynb.

### 4.8 Experiment 8

In this experiment Uni-bi-trigrams of the words of the data obtained after iteration 4 of data cleaning were experimented with minimum n-gram occurrence frequencies ranging from 1 to 5.

Following table shows the F1-scores obtained when uni-bi-trigrams of the tf-idf vectors were used for making the vectors. The *f* in the following table denotes the minimum occurrence frequency of the n-gram and *KC* denotes kernel crashed due to lack of resources.

Table 9. F1-Scores for uni-bi-trigrams of tf-idf vectors with various n-gram frequencies on data obtained after iteration 4 of data cleaning.

| F1-Score | SVM | KNN | Decision Tree | Gaussian Naïve Bayes | Multinomial Naïve Bayes | Logistic Regression | Random Forest |
|---|---|---|---|---|---|---|---|
| f = 1 | 0.6759 | 0.6505 | 0.5845 | KC | KC | 0.6785 | 0.6602 |
| f = 2 | 0.6687 | 0.5819 | 0.5786 | 0.5048 | 0.6692 | 0.6806 | 0.6699 |
| f = 3 | 0.6703 | 0.5655 | 0.5745 | 0.4851 | 0.6762 | **0.6830** | 0.6681 |
| f = 5 | 0.6668 | 0.5033 | 0.5797 | 0.4678 | 0.6689 | 0.6784 | 0.6692 |

The results of other vectorization techniques on uni-bi-trigrams and code can be viewed at https://github.com/sc19gs/Hinglish/blob/master/Experiment%208.ipynb.

### 4.9 Experiment 9

In this experiment ensemble voting classifier, an ensemble of SVM, Logistic Regression and Random Forests was used for making the models. Two variants of the ensemble voting – hard and soft were considered for experiments. The count vectorizer, One Hot Binarizer and Tf-Idf vectorizer were used as the vectorization techniques for creating the vectors of the data. Two settings of these vectorizers were explored which resulted in best results in the previous experiments conducted. First setting was taking Unigrams of words considering minimum occurrence frequency of n-grams as 1 for creating the vectors. Second setting was taking Uni-Bi-Trigrams of words considering minimum occurrence frequency of n-grams as 2 for creating the vectors. This classifier was experimented on data obtained after iteration 4 and iteration 5 of data cleaning.

Table 10. F1-Scores for Experiment 9 on data cleaned using Iteration 4.

| F1-score | Count Vectorizer | | One Hot Binarizer | | Tf-Idf Vectorizer | |
|---|---|---|---|---|---|---|
| | Uni | Uni-bi-tri | Uni | Uni-bi-tri | Uni | Uni-bi-tri |
| Vote (Hard) | 0.6677 | 0.6640 | 0.6613 | 0.6595 | 0.6837 | 0.6816 |
| Vote (Soft) | 0.6774 | 0.6726 | 0.6751 | 0.6651 | 0.6850 | **0.6852** |

Table 11. F1-Scores for Experiment 9 on data cleaned using Iteration 5.

| F1-score | Count Vectorizer | | One Hot Binarizer | | Tf-Idf Vectorizer | |
|---|---|---|---|---|---|---|
| | Uni | Uni-bi-tri | Uni | Uni-bi-tri | Uni | Uni-bi-tri |
| Vote (Hard) | 0.6635 | 0.6575 | 0.6666 | 0.6614 | 0.6745 | 0.6820 |
| Vote (Soft) | 0.6760 | 0.6742 | 0.6739 | 0.6735 | **0.6824** | 0.6823 |

The experiment results and code can be viewed at https://github.com/sc19gs/Hinglish/blob/master/Experiment%209.ipynb.

## 4.10 Experiment 10

In this experiment *n estimators* parameter for the random forests in ensemble voting (soft) classifier was experimented with three more values which were 750, 900 and 1250.

These three classifiers were used on data obtained after iteration 4 of data cleaning and Tf-Idf Vectorizer was used as the vectorization technique where only unigrams and uni-bi-trigrams were considered with their different minimum occurrence frequencies.

Following table shows the F1-scores recorded for this experiment. The *f* in the following table denotes the minimum occurrence frequency of the n-grams.

Table 12. F1-Scores for Experiment 10.

| F1-score | N estimator = 750 | | N estimator = 900 | | N estimator = 1250 | |
| --- | --- | --- | --- | --- | --- | --- |
| | Uni | Uni-bi-tri | Uni | Uni-bi-tri | Uni | Uni-bi-tri |
| f = 1 | 0.6837 | 0.6823 | 0.6847 | 0.6823 | 0.6850 | 0.6806 |
| f = 2 | **0.6907** | 0.6842 | 0.6901 | 0.6856 | 0.6881 | 0.6864 |
| f = 3 | 0.6905 | 0.6867 | 0.6885 | 0.6880 | 0.6896 | 0.6877 |

The experiment results and code can be viewed at https://github.com/sc19gs/Hinglish/blob/master/Experiment%2010.ipynb.

## 5 Conclusion

It was observed that the **Ensemble Voting (soft)** classifier achieved the best F1-score of **0.6907** for classifying the Hinglish code-mixed data from all the experiments conducted above. It was an ensemble of SVM, Logistic Regression and Random Forest where 750 was provided as the values for *n estimators* parameter for the Random Forest and probability parameter for the SVM was set to *true*. The data was transformed into vectors of numbers using the **Tf-Idf** vectorizer taking **Unigrams** of words and considering their minimum occurrence frequency as **2** for making the vectors.

The machine learning algorithm which was able to get the best results was Logistic Regression classifier where data provided was transformed using the Tf-Idf vectorizer taking Uni-Bi-Trigrams of words where minimum n-gram occurrence frequency was considered to be 3. Logistic Regression model got the best results in 6 out of all the experiments conducted. It was followed by Random Forest, SVM, Multinomial Naïve Bayes, KNN, Decision Tree and Gaussian Naïve Bayes in order of ranking as observed from the experiments conducted.

The best results were obtained with the Tf-Idf vectorizer as the vectorization technique. In 8 out of all the experiments conducted Tf-Idf vectorizer gave the best results. The settings for the vectorizer which resulted in the best results was observed to be taking Unigrams or Uni-Bi-Trigrams considering the minimum n-gram occurrence frequency as 2 or 3 for making the vectors. The worst results for the classifiers were obtained using just the Trigrams. So, it was observed that maximum information can mined using Unigrams, Bigrams and Trigrams collectively and the least could be mined using just the Trigrams.

Normalization of Hindi words using a spell-checking dictionary increased the performance of the classifiers significantly. Converting the emoticons into keywords decreased the performance of the classifiers which may be due to the fact that emoticons used by the people in the text could be sarcastic in the context. So, maybe it could not convey the sentiment polarity of the tweets.

## 6   Future Work

More experiments can be conducted with and without using the custom stopword list to analyze the significance of using the custom stopword list and making improvements in it.

Much work is needed to normalize the spelling variations of the Hindi words which can improve the efficiency of the models. Some work also is needed to normalize the spelling mistakes done in writing the English words.

POS tagger for the Hindi words can be created that can tag the Part of Speech for Hindi words. A better POS tagger for English words can be created which can identify the difference between the English words and words written from other languages in the English script. A lemmatizer for the Hindi words can be created that can lemmatize the Hindi words in their basic form. A list of stopwords for the Hindi words can be created.

Vairious neural network architectures such as MLP, CNN, RNN, GRU, Bi-RNN, LSTM, BERT can be experimented. Similarly, many other variations in the data transformation techniques can be experimented which can be Count Vectorizer, Tf-Idf Vectorizer, One Hot Binarizer, Word2Vec embeddings, Doc2vec embeddings, Fasttext embeddings and Glove embeddings.